\documentclass[10pt,twocolumn,letterpaper]{article}

\usepackage{cvpr}
\usepackage{times}
\usepackage{epsfig}
\usepackage{graphicx}
\usepackage{amsmath}
\usepackage{amssymb}
\usepackage{subcaption}
\usepackage{authblk}
\usepackage{array}

\makeatletter
\renewcommand\AB@affilsepx{, \protect\Affilfont}
\makeatother

\usepackage[pagebackref=true,breaklinks=true,letterpaper=true,colorlinks,bookmarks=false]{hyperref}

\cvprfinalcopy 

\def\cvprPaperID{10} 

\ifcvprfinal\fi

\begin{document}

\newcommand\Mark[1]{\textsuperscript#1}

\author[1]{Marcel Sheeny}
\author[1]{Andrew Wallace}
\author[2]{Mehryar Emambakhsh}
\author[1]{Sen Wang}
\author[3]{Barry Connor}
\affil[1]{Heriot-Watt University}
\affil[2]{Cortexica Vision Systems}
\affil[3]{Thales UK}
\renewcommand\Authands{ and }

\title{POL-LWIR Vehicle Detection: Convolutional Neural Networks Meet Polarised Infrared Sensors}

\twocolumn[{%
\renewcommand\twocolumn[1][]{#1}%
\maketitle

\begin{center}
    \centering
    \includegraphics[width=1\textwidth]{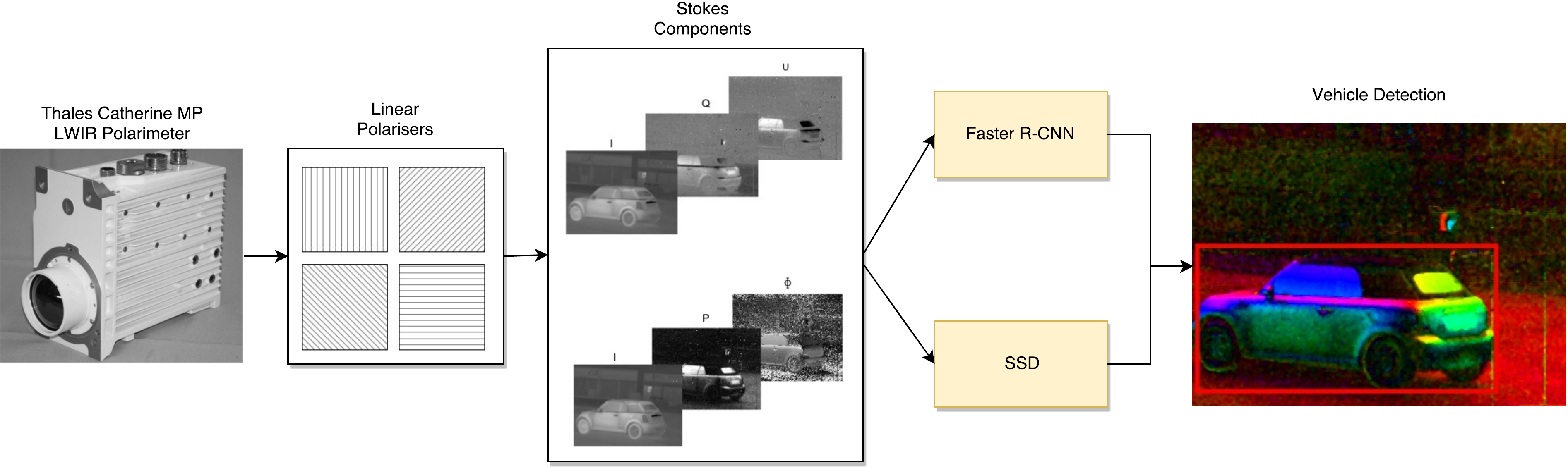}
    \captionof{figure}{POL-LWIR Vehicle detection. This paper uses the Thales Catherine MP LWIR sensor, which is based on long wave polarised infrared technology. It contains 4 linear polarisers ($0^o$, $45^o$, $90^o$, $135^o$). From the linear polarisers we can compute the Stokes components $I$,$Q$,$U$,$P$ and $\phi$.   Two configurations are created ($I$,$Q$,$U$ and $I$,$P$,$\phi$), which are passed to 2 types of neural networks: Faster R-CNN \cite{ren2017faster} and SSD \cite{liu2016ssd}. The networks are trained to detect vehicles in both day and night conditions.}\label{fig:pol-ir}
\end{center}%
}]



\begin{abstract}
For vehicle autonomy, driver assistance and situational awareness, it is necessary to operate at day and night,
and in all weather conditions.
In particular, long wave infrared (LWIR) sensors that receive predominantly emitted radiation have the capability to operate at night as well as during the day.
In this work, we employ a polarised LWIR (POL-LWIR) camera to acquire data from a mobile vehicle, to compare and contrast four different convolutional neural network (CNN) configurations to detect other vehicles in video sequences.
We evaluate two distinct and promising approaches, two-stage detection (Faster-RCNN) and one-stage detection (SSD), in
four different configurations.
We also employ two different image decompositions: the first based on the polarisation ellipse and the second on the Stokes parameters themselves.
To evaluate our approach, the experimental trials were quantified by mean average precision (mAP) and processing time, showing a clear trade-off between the two factors. 
For example, the best mAP result of 80.94 \% was achieved using Faster-RCNN, but at a frame rate of 6.4 fps.
In contrast, MobileNet SSD achieved only 64.51 \% mAP, but at 53.4 fps.
\end{abstract}

\section{Introduction}

The future of autonomous cars is still uncertain, but impressive new results are being achieved with most car manufacturers promising level 4 autonomy by 2020. 
A necessary capability for autonomy is sensory perception,
but the vast majority of research is based on publicly available video benchmarks like KITTI \cite{geiger2013vision} and CityScapes \cite{cordts2016cityscapes}.
These datasets are acquired during daytime, in good weather conditions, using video cameras.
For full autonomy and situational awareness in all weather conditions, sensors and perceptual algorithms workable continuously in 24 hours capability are required.
Infrared sensors are capable of sensing beyond the visible spectrum and are robust to falling illumination.

The majority of previous works, e.g. \cite{lin2015ATR_IR} \cite {khan2014ATR}, have used IR sensors in a military or surveillance context to detect \emph{hot} objects, especially during night.
From the perspective of a commercial road vehicle, IR sensors have also been used at night to detect other actors such as vehicles and pedestrians \cite{kwakIRPed2017}.
However, it has long been recognised that additional analysis of polarisation state,
governed by the material refractive index, surface orientation and angle of observation,
can lead to better discrimination.
False colour representations of the polarimetric data can visually reveal hidden targets \cite{wong2012novel} and to some extent, 3D structure \cite{lavigne2011target}. 
By exploiting knowledge about the dependence of polarisation state on surface and viewing angle, and the
fixed vehicle to road geometry, Connor \textit{et al.} \cite{connor2011scene} constructed a road segmentation
algorithm.
Bartlett \textit{et al.} \cite{bartlett2011anomaly} used polarisation to improve anomaly detection.
Using a nearest neighbour detection based on Euclidean distance, foreground vs background clustering is performed. 
The points whose distance from background components are higher than a threshold are labeled anomalous.
Romano \textit{et al.} \cite{romano2012day} also used polarisation for anomaly detection.
A data cube was formed from the primary ($i_0$,$i_{90}$) imagery and the sample covariance within a local (sliding) window is compared to the sample covariance of the entire image.
They found this method is better at discriminating between target and background  pixels at different times of the day than using Stokes components.

With its dramatic increase in popularity, there have been a number of recent studies based on deep neural networks to recognise objects in IR images. 
For example, Rodger \textit{et al.} \cite{rodger2016classifying} used CNNs to classify pedestrians, vehicles, helicopters, airplanes and drones using a LWIR sensor.
Abbott \textit{et al.} \cite{abbott2017deep} used the YOLO method \cite{redmon2016you} and transfer learning from a high resolution IR to a lower resolution IR (LWIR) dataset to detect vehicles and pedestrians.
Lie \textit{et al.} \cite{liu2016multispectral} used the KAIST dataset \cite{hwang2015multispectral} to fuse RGB and thermal information in a Faster R-CNN architecture, again to detect pedestrians.
Gundogu et al. \cite{Gundogdu2017ATR} combined a CNN detection stage with a long term correlation tracker to detect
tanks in cluttered backgrounds.
However, all of these studies only used intensity data.

Dickson \textit{et al.} \cite{dickson2015long} exploited a polarised LWIR sensor to detect vehicles in both still images and to a lesser extent, video sequences.
In rural settings, LWIR emissions from man-made objects appear more strongly polarised than
vegetation. However in urban settings, most of the environment is man-made.
Therefore, in their work on vehicle recognition \cite{dickson2015long}, the key observation was that although
there are many other man-made structures in urban scenes, there is a distinct, differential spatial arrangement of surface signatures in LWIR polarimetric images of vehicles due to their
regular structure and size.
This leads to a regular pattern of pixel clusters in a 2D space encoding the degree and angle of polarisation.

In this paper, our contribution is to evaluate the effectiveness of CNNs to detect vehicles in polarised LWIR data.
We evaluate the two main research directions in deep learning for object detection: two-stage detection, which first proposes the bounding boxes then performs classification in each bounding box based on Faster R-CNN \cite{ren2017faster}, and one-stage detection which detects and classifies in a single network,
based on Single-Shot Multibox Detection (SSD) \cite{liu2016ssd}.
To the best of our knowledge, this is the first paper to exploit the use of polarised infrared together with neural networks for object detection.
  
\section{Methodology}

\subsection{Sensing and the Polarisation Parameters}
\label{sec:polir}

Figure~\ref{fig:pol-ir} illustrates a schematic diagram of our approach.
We use the Thales Catherine MP LWIR Polarimeter \cite{dickson2015long}, operating in the range of
$8\mu m$ to $12\mu m$ to record video images.
Each pixel of a $ 320 \times 256 $ image frame has $2 \times 2$ sensing sites that contains linear polarisers oriented at $0$, $45$, $90$ and $135$ degrees.
The data capture rate is 100 frames per second (fps).
Our dataset was collected in Glasgow, UK on $14^{th}$ and $15^{th}$ of March, 2013.
Seven sequences are recorded and the bounding boxes of the vehicles are annotated, from which we use 4 sequences for training and 3 for testing.
In total, we have 10659 annotated frames for training and 4453 annotated frames for testing.

The polarisation  state  of  the emitted LWIR radiation can be expressed in terms of the Stokes vector,
$\{I,Q,U,V\}$ \cite{azzam1977ellipsometry}:
The $I$ component measures the total  intensity;  the $Q$ and $U$ components describe the radiation polarised in the horizontal direction and in a plane rotated $45^0$ from the horizontal direction, respectively; the $V$ component describes the amount of right-circularly polarised radiation.
To measure the $V$ component we require an additional quarter wave-plate. As a result we can only measure $I,Q$ and $U$.
Therefore, with respect to the measured intensities at each pixel site, we deduce that

\begin{gather}
I = \frac{1}{2}(i_{0} + i_{45} + i_{90} + i_{135})\\
Q = i_{0} - i_{90}\\
U = i_{45} - i_{135}
\end{gather}

\noindent The degree of linear polarisation, $P$, and the angle of polarisation, $\phi$, can also be calculated as follows,

\begin{gather}
P = \frac{\sqrt{Q^2 + U^2}}{I} \\
\phi = \frac{1}{2}tan^{-1}\Bigg(\frac{U}{Q}\Bigg)
\end{gather}

\subsection{Faster-RCNN}\label{sec:faster}

The Faster R-CNN method \cite{ren2017faster} relies on a two-stage object detection procedure.
First, a sub-network is used to propose the bounding boxes; Second, a separate sub-network
is used to classify objects within each bounding box. 
The idea of Faster R-CNN evolved from R-CNN \cite{girshick2014rich}, which proposes several bounding boxes based on the selective search algorithm \cite{uijlings2013selective}.
Selective search applies a segmentation algorithm in many stages to under and over segment the image.
Bounding boxes are proposed at each region segment; these are inputs to a CNN to classify the type of object.
This CNN can be chosen from popular successful architectures, such as VGG \cite{Simonyan14c},
InceptionNet \cite{szegedy2015going} or ResNet \cite{he2016deep}.

Since selective search usually outputs large number of regions ($\sim$2000 regions), it is computationally expensive.
Fast R-CNN \cite{girshick2015fast} reduces this complexity by running a CNN over the whole image.
Proposed regions are transformed to the last feature map before the fully connected layers and the regions in the feature map are classified in a simple neural network, resulting in a significant computational complexity reduction. 
Despite such improvement, the authors realised that the selective search is indeed a bottleneck preventing faster execution of the overall algorithm.
As a solution, Faster R-CNN \cite{ren2017faster} created a network to learn how to generate bounding boxes.
This pipeline is also known as region proposal network (RPN).
The RPN creates a grid in the original image which anchors the bounding box annotations to the map.
Using the previous annotations of the bounding box in the original image, the RPN learns how to propose bounding boxes. RPN can also reduce the number of proposals compared to selective search.
Replacing selective search with the RPN improved both speed and accuracy (using the PASCAL VOC 2012 dataset).

Since it is easy to plug any CNN into the Faster R-CNN method, we decided to use ResNet-50 and ResNet-101 \cite{he2016deep}, where the number attached to the name of the network relates the number of layers used.
ResNet uses residual layers that are CNNs with “shortcut connections”.
Those connections skip the current layer and the skipped output is added to the output after the convolution is applied.
ResNet has a trade-off between accuracy and depth of the network: the smaller the network is, the faster it performs.

\subsection{Single Shot Multi-box Detection}\label{sec:ssd}

Single Shot Multi-box Detection (SSD) \cite{liu2016ssd} uses one-stage detection,
in which the output of a single network is a set of bounding boxes with the respective classes.
This is different from Faster R-CNN which has two stages, the region proposal and the classification stages.

The use of one-stage detection attracted the attention of many researchers in the field.
Sermane \textit{et al.} \cite{sermanet2014overfeat} used a CNN over the whole image, where each cell of the last feature map corresponds to a region in the original image.
The regions are always uniformly distributed over the image, which constitutes a major disadvantage as there is
no {\emph{a priori}} reason why this should be the case.
The YOLO network \cite{redmon2016you} created a CNN which is trained using anchor boxes from the annotation (similar to RPN from Faster R-CNN) to output the region location plus its classification.
SSD is similar to YOLO in the sense that it uses a CNN to output the region's location and its classification result.
However, SSD generates the output at several scales of the feature map produced by the convolution.
The output maps are based on a grid of anchor boxes, as with Faster R-CNN. 
All the results at several scales are then combined, followed by a non-maximum suppression step to remove
multiple detections of the same object.

The architecture of SSD is based on convolutional stages of other networks,
such as VGG \cite{Simonyan14c} and InceptionNet \cite{szegedy2015going} and MobileNet \cite{howard2017mobilenets}.
InceptionNet (GoogleNet) was the winner of the ImageNet 2014 competition for image classification.
This network applies convolutions of different sizes ( $1 \times 1$, $3 \times 3$, $5 \times 5$) in the same layer and concatenates them into a single feature map.
These convolutions are called "Inception modules", and when done several times create a deep network of inception modules. 
This showed that applying several convolutions to the same feature map can retrieve more robust features.
MobileNet \cite{howard2017mobilenets} was designed to be a fast and small CNN to run on low powered devices.
It is a convolutional layer approximator;
Instead of applying $N \times M \times C$ convolutions, it first applies a $1 \times 1 \times C$ convolution to reduce the size of the input to $W \times H \times 1$, then applies a $N \times M \times 1$ convolution (where $N$ and $M$ are the convolution mask sizes, $W$ and $H$ are width and height of the image and $C$ is the number of channels).
This strategy reduces the number of weights to be learned by the network and the complexity of the convolution.
SSD with MobileNet is used in our experiments to evaluate how a small network can learn the polarised infrared features.

\begin{figure*}[h!]
\centering
\begin{subfigure}[t]{0.5\textwidth}
\centering
\includegraphics[height=1.8in]{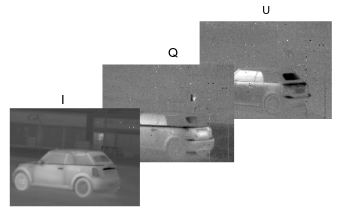}
   \caption{$I,Q,U$ configuration.}
\label{fig:iqu}
\end{subfigure}%
~
\begin{subfigure}[t]{0.5\textwidth}
\centering
\includegraphics[height=1.8in]{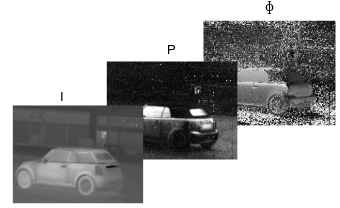}
   \caption{$I,P,\phi$ configuration.}
\label{fig:ipphi}
\end{subfigure}
\caption{Visualisation of each configuration based on the Stokes components.}
\end{figure*}

\subsection{Experiments, Training and Evaluation}

We have evaluated two configurations of the measured polarised image data to train and test
our several CNN architectures.
The first configuration uses the $I,Q,U$ parameters as the input image planes. 
In Figure~\ref{fig:iqu}, we can visualise the image of each component $I,Q,U$.
Again, in Figure~\ref{fig:ipphi} we can visualise the image of each component $I,P,\phi$.
We use four configurations of neural network for our experiments.
\pagebreak

\begin{enumerate}
\item 
The SSD network using the InceptionV2 network \cite{szegedy2015going} to extract features.
\item
The SSD network using the MobileNet network \cite{howard2017mobilenets} to extract features.
\item
Faster R-CNN using ResNet-50 \cite{he2016deep} to extract features.
\item
Faster R-CNN using ResNet-101 \cite{he2016deep} to extract features.
\end{enumerate}

We trained our 4 different configurations on both ($I,Q,U)$ and $(I,P,\phi$) data, and
for comparison with previous work that has applied CNNs to intensity data alone,
on the $I$ data in isolation.
The networks are trained using a i7-7700HQ, 32 GB ram, NVIDIA Titan X and developed using the Tensorflow Object Detection API \cite{abadi2016tensorflow}.
The network weights for both SSD and Faster R-CNN are initialised from the MS-COCO object
detection dataset \cite{lin2014microsoft}.
The parameters for the Faster R-CNN networks are: batch size $1$, learning rate $0.0003$, momentum $0.9$.
The parameters for the SSD networks were: batch size $24$, learning rate $0.004$, momentum $0.9$.
Eq. \ref{eq:sgd} shows the gradient descent formula.

\begin{equation}\label{eq:sgd}
W_{t+1} = W_{t} - \alpha \nabla f(x;W) + \eta \Delta W 
\end{equation}

\noindent
where $\eta$ is the momentum, $\alpha$ is the learning rate, $t$ is the time current time step, $W$ is all weights of the network and $\nabla f(x;W)$ is the derivative of the function that represents the network and $x$ is our dataset.

\begin{figure*}[t!]
    \centering
    \begin{subfigure}[t]{0.33\textwidth}
        \centering
        \includegraphics[height=2.0in]{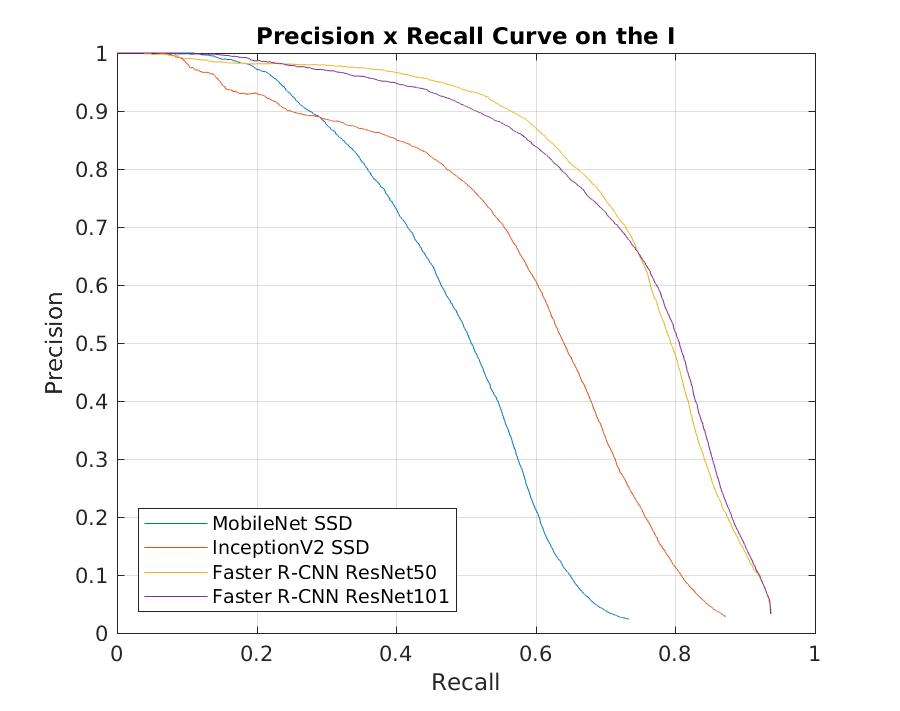}
        \caption{Precision-Recall curve for the $I$ configuration.}\label{fig:pr_i}
    \end{subfigure}%
    ~ 
    \centering
    \begin{subfigure}[t]{0.33\textwidth}
        \centering
        \includegraphics[height=2.0in]{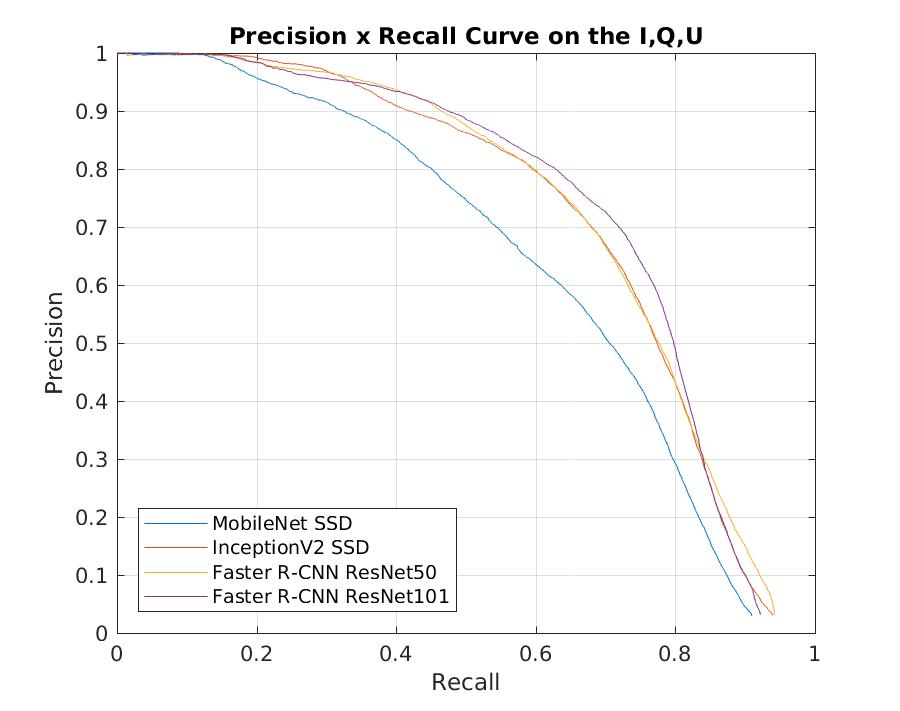}
        \caption{Precision-Recall curve for the $I,Q,U$ configuration.}\label{fig:pr_iqu}
    \end{subfigure}%
    ~ 
    \begin{subfigure}[t]{0.33\textwidth}
        \centering
        \includegraphics[height=2.0in]{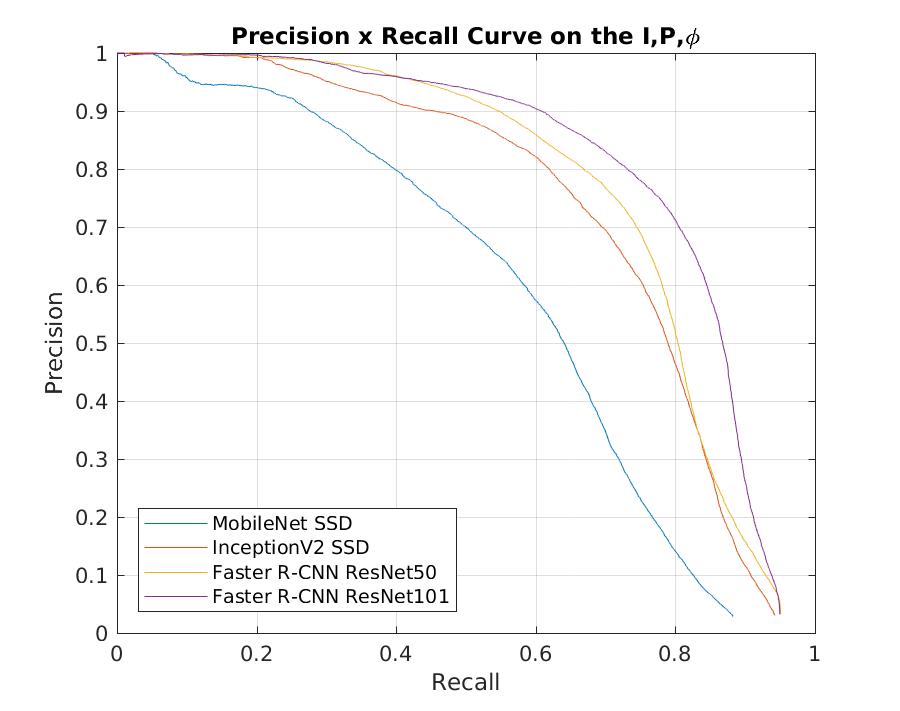}
        \caption{Precision-Recall curve for the $I,P,\phi$ configuration.}\label{fig:pr_ipphi}
    \end{subfigure}
    \caption{Precision-Recall curves}
\end{figure*}

The evaluation metrics used are mean average precision (mAP) and processing time in frames per second (fps).
The mAP classifies correct detection when Intersection over Union (IoU), $ > 0.5$, which follows the PASCAL VOC
protocol \cite{everingham2015pascal}. 
(The KITTI protocol \cite{geiger2013vision} uses $IoU > 0.7$ for vehicles. However, since unlike KITII, our annotations are not pixel-level, we followed the PASCAL criteria.)
To compute the fps we compute the average time over 100 frames.
Tables \ref{tab:results} and \ref{tab:speed} show the results for each configuration.
Precision-recall curves are also generated for each results and can be visualised
in Figure~\ref{fig:pr_i} for $I$ alone, Figure~\ref{fig:pr_iqu} for $I,Q,U$ and Figure~\ref{fig:pr_ipphi} for $I,P,\phi$.

\setlength\tabcolsep{1.5pt}
\begin{table}[]
\centering
\caption{Results for each configuration}\label{tab:results}
\begin{tabular}{l|ccc}
                       & mAP [I] & mAP [I,Q,U] & mAP [I,P,$\phi$]   \\ \hline
MobileNet SSD          & 48.50 \%  & 64.51 \% &58.56 \%     \\         
InceptionV2 SSD        & 59.79 \%  & 72.17 \%&73.24 \%      \\      
Faster R-CNN Resnet-50 & 75.63 \% & 72.82 \% &76.43 \%      \\
Faster R-CNN Resnet-101 & 75.21 \% & 73.67  \%   &\textbf{80.94} \%        
\end{tabular}
\end{table}

\begin{table}[]
\centering
\caption{Computational speed (fps) for each network configuration.}\label{tab:results_speed}\label{tab:speed}
\begin{tabular}{l|c}
                       & fps \\ \hline
MobileNet SSD          & \textbf{53.4}    \\
InceptionV2 SSD        & 37.2     \\
Faster R-CNN Resnet-50 &  7.8    \\
Faster R-CNN Resnet-101 & 6.4
\end{tabular}
\end{table}

\begin{figure*}[h!]
\begin{center}
\includegraphics[width=0.8\textwidth]{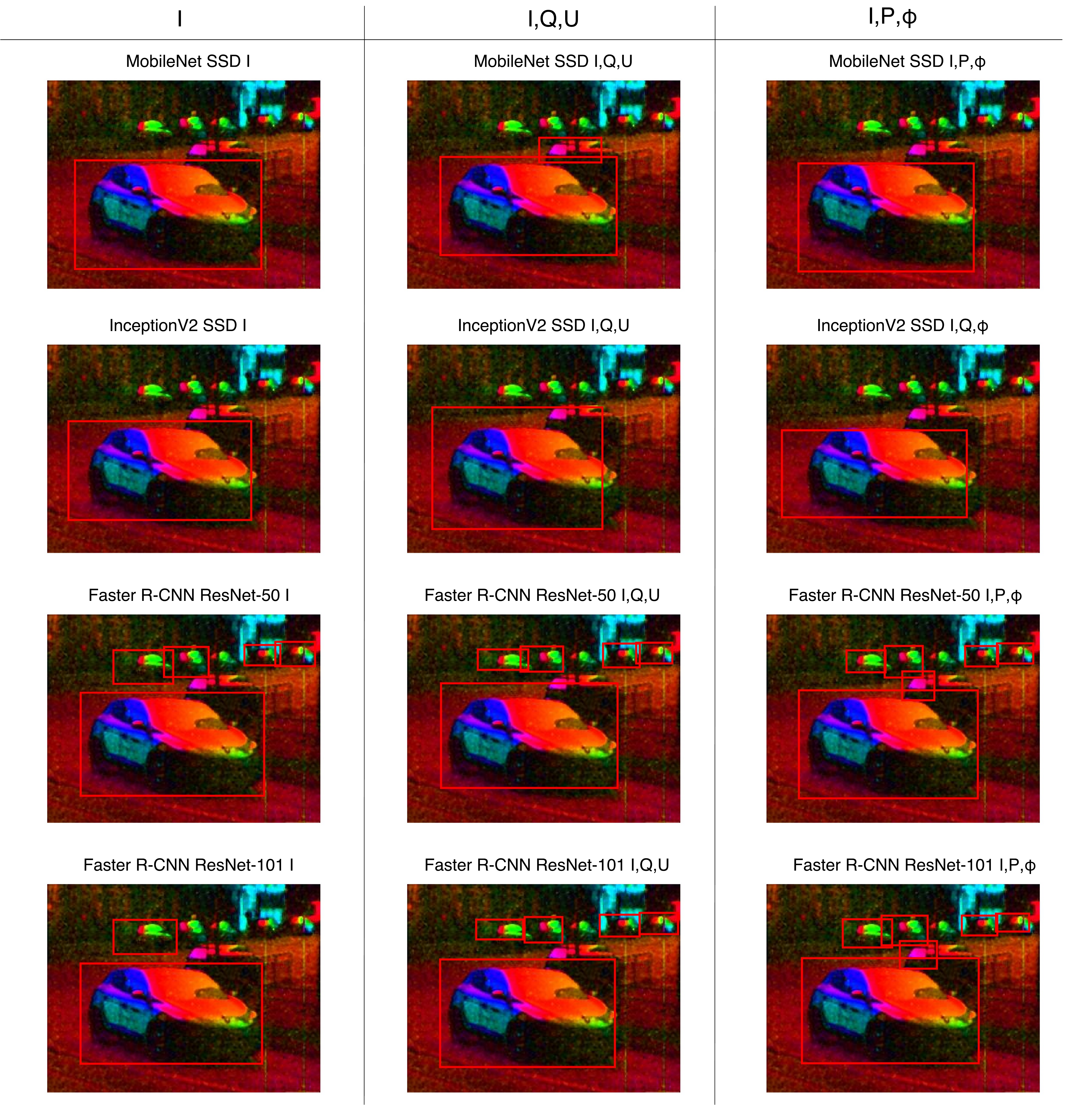}
\end{center}
   \caption{Qualitative results for each configuration with threshold at 0.7.}
\label{fig:quali}
\end{figure*}

Qualitative examples of images can be seen in Figure~\ref{fig:quali}.
This uses a pseudo-colour display, converting $P$ and $\phi$ to HSV colour space.
Based on qualitative results we can see that the main problem of the SSD lies with small objects.
The network needs to learn the location and features at the same time, which affects the detection of small objects.

Considering the results of Table~\ref{tab:results} and the precision-recall curves, a key question
is whether the use of polarised data improves detection when compared to the intensity data alone,
as used by most previous authors.
We believe that it is indeed the case, particularly for the SSD examples,
although we qualify that statement by noting that the difference in the R-CNN results
is less definitive, and that the dataset is limited and many more trials are needed
for full statistical verification.
In general, for all datasets, Faster R-CNN performs better for this limited trial, as measured by mAP, although this comes with the penalty of a much slower frame rate.
However, this latter result is consistent with the published results on video sequences, where splitting the
tasks of region proposal and classification arguably makes the network more robust in learning
directly the object features.
In contrast, SSD needs to learn the localisation together with classification, and hence
both location and object characteristics influence the weights of the network which can degrade performance.
Nevertheless, in our trial, SSD-InceptionV2 achieves similar performance to Faster R-CNN ResNet-50,
at a much increased speed, since it just needs one CNN for both region proposal and classification.

Comparing $I,Q,U$ and $I,P,\phi$, the best result is obtained with the $I,P,\phi$ parametrisation. Although
the differences are not proven as significant, such that much more extensive characterisation is required.
At this stage, given the complexity of these neural networks, it is hard to define what type of feature
is being learned in each case, although from Dickson \textit{et al.} \cite{dickson2015long} the authors claim that material,
shape and surface and viewing angles influence the underlying polarisation patterns. 
As a specific example, one can see that the $I,P,\phi$ configuration does detect an occluded car that the $I,Q,U$ does not in the Faster R-CNN example. However, although this occurs more often than the converse, much better understanding of the network and more extensive trials are necessary to draw reliable conclusions.
For a necessary perception by an autonomous car, computational time is clearly quite a crucial factor.
As expected, the one-stage SSD architectures shows higher frame rates.
MobileNet SSD is the fastest and can process on average at 53.4 fps, but it has the lowest mAP.\vfill\break

\section{Conclusions}

In this paper, we evaluated and compared a series of different CNN architectures for vehicle detection
in polarised long wave infrared image sequences, using two different image decompositions.
We showed that the use of polarised infrared data was effective for vehicle detection, and appeared to
perform better when CNNs are used for detection in infrared intensity data alone, confirmed also by previous researches.
Faster R-CNN based networks achieved better results in terms of detection accuracy, splitting the tasks of
region proposal and classification to  make the network more robust. 
However, it should be mentioned that improving the accuracy of one-stage detection network is quite an active field of research, providing much higher frame rates.
We could reach no firm conclusion on which image decomposition was preferable, although anecdotally the $\{I, P, \phi\}$ parametrisation is both more intuitive in describing the polarisation ellipse and achieved the best overall result
with Faster R-CNN ResNet 101. Our detection rates are similar to those of KITTI dataset for vehicle detection from simple video data in daylight using the same networks. Overall, our work shows that polarising LWIR data is a relatively robust
option for day and night operation.

\section*{Acknowledgements}

We acknowledge the support of the Engineering and Physical Research Council, grant references EP/N012402/1 
(TASCC: Pervasive low-TeraHz and Video Sensing for Car Autonomy and Driver Assistance (PATH CAD)) and
EP/G037523/1 (Optics and Photonics Technologies) conducted with and supported by Thales UK. 

Thanks to NVIDIA for the donation of the TITAN X GPU.





{\small
\bibliographystyle{ieee}

}

\end{document}